# Lumen Shape Reconstruction using a Soft Robotic Balloon Catheter and Electrical Impedance Tomography

James Avery, Mark Runciman, Cristina Fiani, Elena Monfort Sanchez, Saina Akhond, Zhuang Liu, Kirill Aristovich and George Mylonas

*Abstract*— Incorrectly sized balloon catheters can lead to increased post-surgical complications, yet even with preoperative imaging, correct selection remains a challenge. With limited feedback during surgery, it is difficult to verify correct deployment. We propose the use of integrated impedance measurements and Electrical Impedance Tomography (EIT) imaging to assess the deformation of the balloon and determine the size and shape of the surrounding lumen. Previous work using single impedance measurements, or pressure data and analytical models, whilst demonstrating high sizing accuracy, have assumed a circular cross section. Here we extend these methods by adding a multitude of electrodes to detect elliptical and occluded lumen and obtain EIT images to localise deformations. Using a 14 Fr (5.3 mm) catheter as an example, numerical simulations were performed to find the optimal electrode configuration of two rings of 8 electrodes spaced 10 mm apart. The simulations predicted that the maximum detectable aspect ratio decreased from 0.9 for a 14mm balloon to 0.5 at 30mm. The sizing and ellipticity detection results were verified experimentally. A prototype robotic balloon catheter was constructed to automatically inflate a compliant balloon while simultaneously recording EIT and pressure data. Data were collected in experiments replicating stenotic vessels with an elliptical and asymmetrical profile, and the widening of a lumen during angioplasty. After calibration, the system was able to correctly localise the occlusion and detect aspect ratios of 0.75. EIT images further localised the occlusion and visualised the dilation of the lumen during balloon inflation.

## I. INTRODUCTION

Balloon catheters are widely employed in minimally invasive surgery to dilate and restore patency to obstructed lumen inside the body. During percutaneous transluminal angioplasty (PTA) inflation of the balloon is used to eliminate blockage caused by a stenosis or plaque in periphery or coronary vessels [1], [2]. They are also used to convey and deploy stents to maintain long term vessel patency, or other prostheses such as valve replacements in Transcatheter Aortic Valve Replacement (TAVR) [3]. Prior to TAVR, Balloon aortic valvuloplasty (BAV) is often performed to dilate the native aorta before implantation [1], [4]. Balloons are typically designed to reach a specific diameter for a given pressure and are carefully selected prior to surgery. Thus, accurate feedback of the shape during inflation is important to ensure correct

* James Avery is an Imperial College Research Fellow. This work was part funded by the NIHR Imperial BRC and the Cancer Research UK Grant C57560/A30036.

J. Avery, M. Runciman, E. Monfort Sanchez, S. Akhond, G. Mylonas are with the Hamlyn Centre, Imperial College London, London, W2 1PF, U.K (e-mail: james.avery@imperial.ac.uk). C. Fiani and K. Aristovich are with the Dept. of Medical Physics and Biomedical Engineering, University College London, London, WC1E 6BT, U.K. Z. Liu is with the Dept. of Bioengineering, Imperial College London, London, W2 1PF.

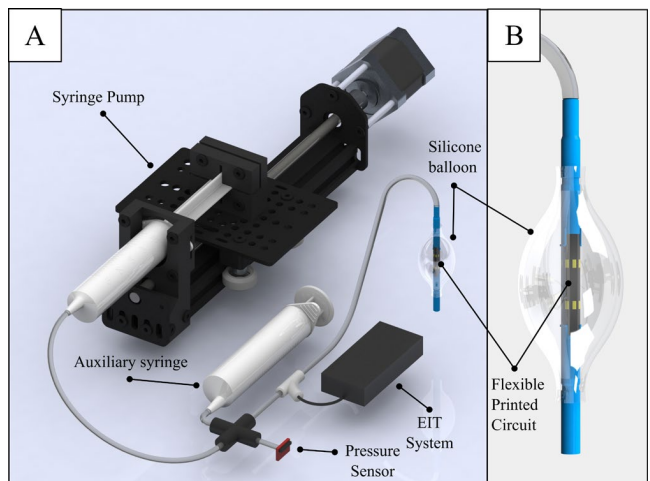

Fig 1. Soft robotic balloon catheter with integrated Electrical Impedance sensing (A) Outer surface of central catheter is modified with thin flexible circuit board; the balloon is then controlled via syringe pump and pressure sensor (B) Detailed view of electrode flexible circuit board.

deployment of the balloon or connected stent and to minimise complications arising from the procedure. During angioplasty the lumen wall can be damaged from high mechanical stresses, and prolonged overpressures have been linked to higher restenosis rates [5].

A critical step in the implantation of the prosthesis is measuring the lumen cross section, to best determine the stent size, reduce patient-prosthetic mismatch and prevent complications. For example, paravalvular leak (PVL) is the main poor cardiac outcome after TAVR, and has been shown to be linked with systematic undersizing of the aortic stent [6], [7]. Oversizing may disrupt the annulus wall or cause a bundle branch block resulting from a conduction disorder [8]. One of the challenges with correctly sizing the aorta is that the cross section is often elliptical, or otherwise non uniform [9]. Presurgical assessments such as 2D Transoesophageal (TOE) or Transthoracic Echocardiography (TTE) measure the aortic cross section assuming a circular cross section using the continuity equation [10]. These have been shown not to determine the size and shape with high accuracy, and thus Multidetector Computed Tomography (MDCT) is recommended [11]. Whilst offering greater accuracy, MDCT cannot be performed during surgery and exposes the patient to side effects of Xray and contrast dyes [4].

Automated control of the balloon inflation using pressure and volume measurements have been shown to improve outcomes in coronary angioplasty [12]. Recently a sizing algorithm has been developed combining robotic pressure control with an analytical inflation model which can aid in accurate annulus sizing, but assumes a circular and consistent cross section [13]. Intravascular optical coherence tomography (IVOCT) images have been proposed as a method of providing balloon diameter feedback during controlled inflation [14], [15]. The images provided are highly accurate but are limited to a single thin cross section, require expensive equipment, and a specialised OCT catheter, which may not be compatible with all balloons with integrated instrument channels.

To prevent air embolisation in the case of a rupture, balloon catheters are inflated with normal saline (0.9% concentration), sometimes mixed with heparin or contrast agents. Taking advantage of the conductivity of the injected saline, successful measurements of the balloon size using electrical impedance methods have also been demonstrated [4], [16]. The functional lumen imaging probe (FLIP) [17] measures impedance between neighbouring pairs of electrodes to quantify cross sectional area (CSA) along the length of the balloon. However, as the ring electrodes cover the complete circumference of the catheter, these devices are not able to detect ellipticity. These methods can be extended to perform Electrical Impedance Tomography (EIT), a technique which reconstructs conductivity distributions within an object from a multitude of impedance measurements. Recently, intravascular EIT imaging using 32 electrodes across the catheter circumference has been proposed for atherosclerosis detection and proximity to the vascular wall [18]. EIT tactile sensors have been developed that leverage the impedance contrast between the electrically conductive saline and insulating polymers, imaging changes within a whole chamber [19], [20] or within microchannels [21], [22].

We propose to extend the impedance sensing methods to create EIT images of the shape of the internal cavity of the balloon catheter. Imaging deformation of the balloon wall has the potential for localising areas of vascular stenosis or measuring the extent of dilation during angioplasty. The EIT balloon catheter concept, Fig. 1, has two rings of electrodes on the outer surface of the central catheter, to minimise the interference with potential guide wire or instrument channels and to avoid the need for complicated stretchable electronics.

The prototype was based on a 14 Fr (5.3 mm) catheter as it is commonly used in TAVR [23], whilst remaining large enough to rapid prototype. Initially, numerical modelling was performed to find the optimum electrode arrangements. Then both the limits of balloon size and ellipticity detection are estimated based on the typical SNR of an EIT system. A prototype catheter was then constructed based on these simulations and results compared in phantom experiments. Finally, a roboticised balloon catheter was built which enabled synchronised recording of volume, pressure and EIT data during inflation. Experiments were then performed to replicate different clinical scenarios: detecting localised deformations *i.e.,* atherosclerosis, detecting lumen ellipticity *i.e.,* TAVR or BAV, and dilation during angioplasty.

## II. NUMERICAL ANALYSIS

First, a simulation study was performed to verify the imaging concept and to optimise the electrode layout. The simulations were performed using EIDORS [24] to solve the EIT Forward problem in models representing a variety of balloon geometries. Finite Element Meshes of *c.* 2million elements were constructed with x10 refinement at the electrodes, Fig. 2A, with homogenous conductivity of 1.6 S/m the same as 0.9% normal saline. The current amplitude was set to 141 μA, based on IEC-60601 safety limits.

### A. Optimal Axial Spacing

The optimal ring spacing is a trade-off between limiting the spread of current and maximising the measurement sensitivity at the balloon wall. A closer spacing reduces the current spread and allows the cross section to be divided into a greater number of independent sectors, and thus better detect elliptical profiles. A wider spacing increases the current at the balloon wall and thus measurement sensitivity $J$ (Vm/S) to changes in these elements. Simulations were performed in an elliptical model with 26 mm major axis and ratio of major to minor axis or *aspect ratio* $f = 0.75$. The spacing between the two rings of electrodes on the catheter $l$ varied from 5 and 40 mm, and measurements were made in 4 overlapping sectors between the major and minor axis of the ellipse $V_{1-4}$. Current was injected straight along the catheter between equivalent electrodes on each ring, and voltages recorded on the neighbouring pair, Fig. 2B. The current density magnitude $CD$ and sensitivity $J$ were calculated and elements within a 2mm slice equidistant from both electrode rings were used in subsequent analysis. The current spread angle $CD_\theta$ was defined as the angle which contained 99% of all elements above half the maximum current in the slice. $J_{wall}$ was defined as the maximum $J$ in the outer elements of the 2mm slice.

For spacings above 15 mm, $CD_\theta$ was greater than 180°, approaching the entire diameter at $l = 30$ mm. Whereas below 15 mm, the current spread angle was localised to 90° or less. This is also reflected in the $J_{wall}$ values which are equal for all four measurements above 30 mm as the current was spread uniformly throughout the balloon and were thus no longer sensitive to local changes in radius. A separation of 10 mm was chosen for all subsequent experiments as it provided the greatest magnitude and spread of $J_{wall}$ and limited current spread to 60°. For a total of 16 electrodes, two rings of eight equally spaced electrodes were chosen as a compromise

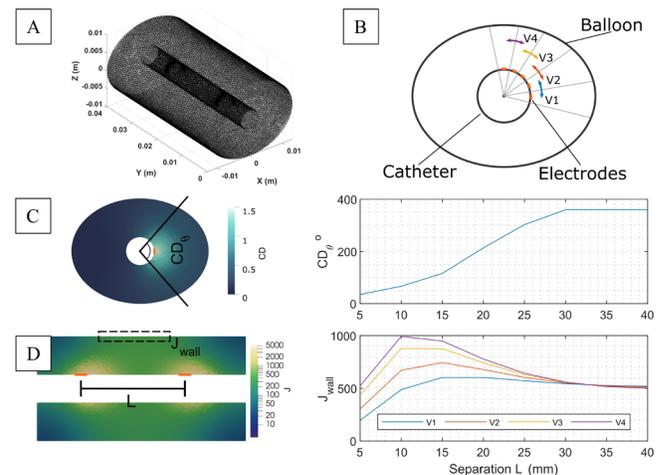

Fig 2. Optimisation of electrode spacing in simulation, (A) Finite element model of electrodes on catheter and elliptical balloon (B) cross section of balloon catheter at first electrode plane with four measurements across different sectors (C) spread of current density $CD_\theta$ with increasing electrode separation $L$ (D) maximum sensitivity at balloon wall $J_{wall}$ with increasing $L$ for each sector

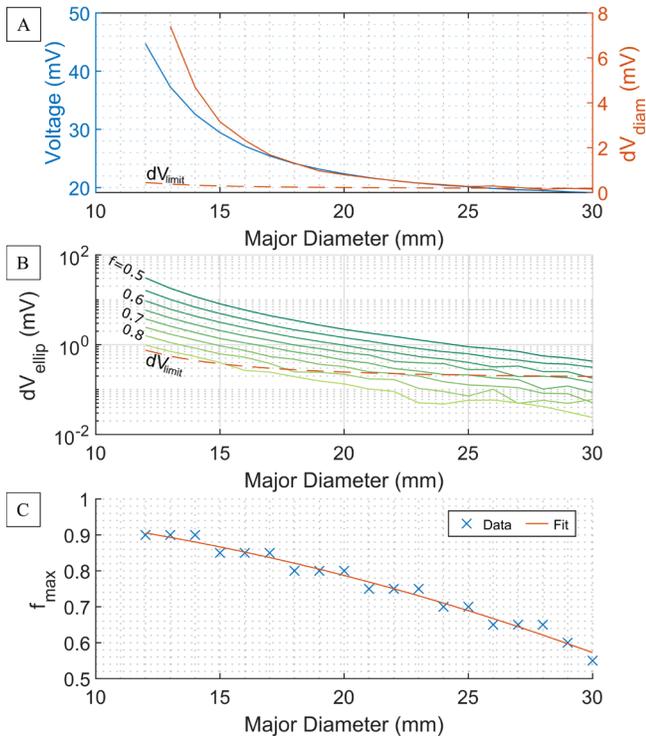

Fig 3. Limits of diameter and ellipticity detection in simulation. Differences greater than $dV_{limit}$ -10 times the noise of 60 dB SNR system - were considered detectable (A) *blue:* voltages in circular models *red:* change in voltage per 1 mm diameter increase $dV_{diam}$ (B) difference between voltages measured on major and minor axis $dV_{ellip}$ for increasing aspect ratios $f$ (C) maximum detectable aspect ratio $f_{max}$ (0.05 increments) and 2nd order fit

between number of independent measurements and hardware and manufacturing limitations.

## B. Balloon Size and Ellipticity Detection Limits

Using this spacing, simulations were performed in a total of 209 meshes with outer diameters of 12 – 30 mm in 1 mm increments and aspect ratios $f$ ranging from 0.5 to 1 in 0.05 increments. Noise was simulated with a SNR of 60 dB (0.1ppm), typical values for an EIT system [20], [25]. The two signals of interest were: the change in voltage for each 1mm increase in diameter for circular balloons $dV_{diam}$, and the difference between measurements at the major and minor axis for elliptical balloons $dV_{ellip}$. For each balloon, a minimum signal amplitude was calculated $dV_{limit}$, equal to 10 times the noise expected from the 60 dB SNR system. Thus a 1 mm increase in diameter $dV_{diam}$ was considered detectable only if $dV_{diam}$ was greater than $dV_{limit}$. Similarly, an aspect ratio was considered detectable if $dV_{ellip}$ was greater than $dV_{limit}$.

The results suggest a balloon size limit of 28 mm diameter as, beyond this, increases in circular CSA were not sufficiently greater than the noise, Fig. 3A. The potential to detect ellipticity was also highly dependent on the major diameter, Fig. 3B. Voltage changes caused by aspect ratios as high as 0.9 were detectable at 14 mm diameter, whereas this decreased to 0.7 when the balloon diameter was 25 mm, Fig. 3C. This suggests that the ability of the device to detect ellipticity is dependent on the expected lumen geometry, with applications in narrower lumen having greater potential to accurately measure ellipticity.

## C. Experimental validation

Phantom experiments were performed, using a prototype constructed from a Flexible Printed Circuit (FPC) and a 3D printed catheter. The FPC had 16 1x2 mm Electroless nickel immersion gold (ENIG) contacts in two rings of eight matching those from the simulations, Fig. 4A. The 14Fr catheter model, Fig. 4B, had a single internal lumen to inflate and deflate the balloons through the two inlet holes, and guides to align the placement of the FPC. EIT measurements were taken using an Eliko Quadra Spectroscopy system [25] with a 16 channel multiplexer. The optimum measurement settings were based on preliminary testing of 3 FPCs in a large saline tank. A frequency of 11 kHz and measurement time of 5 ms was chosen for all subsequent experiments as it offered the highest overall SNR (72.3 dB) and the lowest interchannel variation (< 5%). A *radial* measurement protocol was used wherein current was injected between each ring using a pair of equivalent electrodes and voltages measured on the neighbouring pair, for a total of 8 measurements.

Calibration data was collected in circular containers, filled with 0.9% saline, Fig. 4C. Data was collected 5 times for diameters ranging from 12.5 to 30 mm, and the average repeatability error across all experiments was 0.49 mV

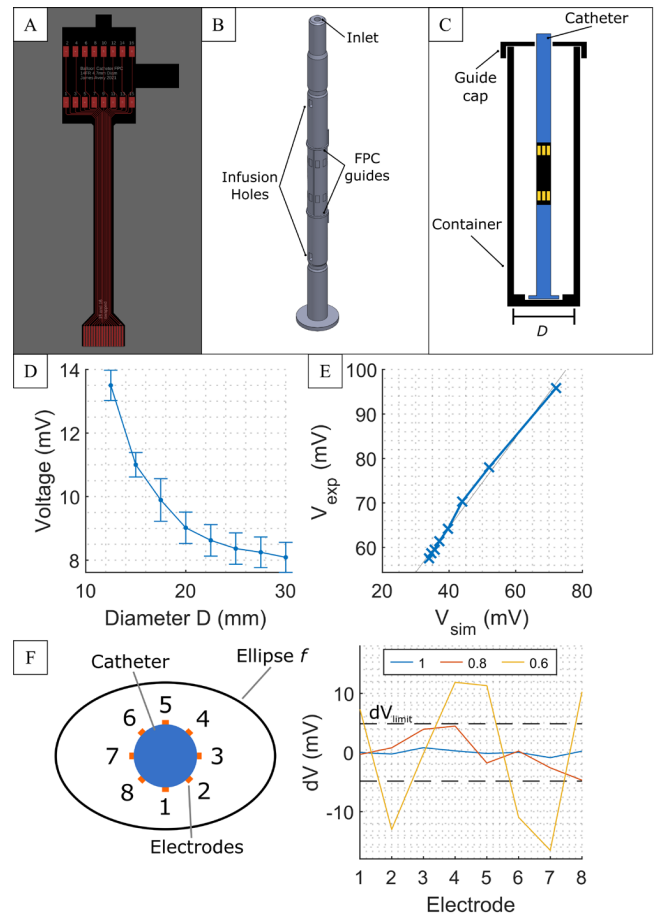

Fig 4. Experimental validation (A) FPC design (B) catheter model guides for FPC and hydraulic inlets for inflation (C) container experiment overview (D) Voltages measured in circular cups with increasing diameter (mean±std) (E) Comparison of simulated $V_{sim}$ and experimental $V_{exp}$ voltages $R^2 = 0.99$ (F) measurement variation for elliptical cross sections $f = [1, 0.8, 0.6]$ for $D = 25$ mm $dV_{limit}$

equivalent to 5.3 %, Fig. 4D. Whilst differences in contact impedance and saline concentration created a mismatch in absolute values, the experimental results closely follow the trend expected from simulations, Fig. 4E, with a $R^2 = 0.99$ linear correlation. Measurements were also collected in elliptical tanks with a major diameter of 25 mm, equal to the largest annulus size for a TAVR valve deployed using a 14 Fr catheter [23]. The results match the prediction from the numerical simulations, Fig. 3C, for a detection limit of 0.75 for $D = 25$ mm. The increase in voltage due to the decreased radius at the minor axis for $f = 0.6$ was clearly distinguishable in the measurements above $dV_{limit}$, Fig. 4F. Whereas voltage changes when $f = 0.8$ were masked by the noise and systematic interchannel variation and were thus below the detection limit $dV_{limit}$. These results demonstrate the principle and verify the simulations, but also highlight the need for careful calibration for accurate ellipticity detection.

## III. METHODS

An automated syringe pump, Fig. 5A, was constructed to control inflation of a compliant balloon, whilst simultaneously recording EIT, pressure and volume data. A second prototype EIT catheter was enclosed by a silicone balloon (Ecoflex 00-20, Smooth-On, USA) consisting of a 35 mm long tube of 7 mm inner diameter and 12 mm outer diameter. The balloon was sealed onto the catheter at both ends with elastic bands that axially stretched the balloon by approximately 66% of its original length. This pre-stretch increased the ratio of radial to axial deformation that occurred when the balloon was inflated, Fig. 5B. The balloon was connected to a syringe pump with a 50 ml supply syringe (50 ml Luer Slip Syringe, BD, USA), filled with 0.9% saline, and air bubbles were removed using an auxiliary syringe. Thus, the saline acted as both the hydraulic actuation fluid and the conductive medium for EIT measurements. The position of the supply syringe piston was controlled by a linear actuator (ACME Lead Screw Linear Actuator, Ooznest, UK) with integrated driver (uStepper S-Lite, uStepper, Denmark), which was interfaced with MATLAB on a control computer.

In each experiment the balloon catheter was placed vertically inside a tubular structure with a given shape and material stiffness. When the volume of the balloon catheter was increased, the outer walls conformed to the shape of the surrounding structure. During each test, a MATLAB script increased the volume of saline within the balloon in ten equal increments from a minimum of 2.21 ml to a phantom dependent maximum value. This minimum value, Fig. 5A, prevented saturation of voltage readings by ensuring saline was present between the electrodes and allowed the identification of air bubbles during the initial setup. Maximum volume values were chosen to prevent damage to the balloon from over-inflation. The maximum volume injected into the balloon at any time during testing was 35.85 ml. At each volume increment, the mean of ten pressure readings was recorded with a pressure sensor (MS580314BA01-00, TE Connectivity, Switzerland) and an EIT measurement frame was taken.

### A. EIT Measurement Setup and Imaging

Impedance measurements were collected using the Eliko Quadra system with the same settings as the previous section. Data were collected using two separate EIT measurement protocols, *radial:* 8 measurements as in the previous section and *full:* a complete EIT protocol for image reconstruction, where current was injected across rings and voltages were measured on all other opposite pairs (x64), also measurements were made on each ring separately using adjacent pairs (2x64) for a total of 192 measurements. This was reduced to 136 after removing measurements on current injecting electrodes. The frame rate was 25 Hz and 1.5 Hz for the *radial* and *full* protocols respectively.

Clinically the lumen size would not be known *a priori*, thus a mesh of the largest expected diameter (30mm) was used to calculate the forward model. EIT reconstructions were performed using two different methods: *Absolute* where the conductivity distribution σ is calculated without any reference frame, and *time difference* where the change in conductivity δσ is reconstructed with respect to a reference data frame, either: simulated voltages of the unobstructed lumen for *Pseudo Time Difference (PTD)*, or voltages recorded at the initial state for conventional *Time Difference (TD)*. Absolute imaging is particularly sensitive to modelling errors [26], however, the simple geometry and large conductivity contrasts make this a suitable application of the algorithm. The inverse problem in EIT is severely ill-posed, and thus it is necessary to apply regularisation techniques to obtain a stable solution [27]. Absolute images were reconstructed in a lower density tetrahedral mesh of 10k elements, using the EIDORS implementation of the regularised Gauss-Newton solver [24] with a NOSER prior [28] with a hyperparameter $\lambda = 0.01$. Time difference images were reconstructed in a 12k element hexahedral mesh and Tikhonov Zeroth order regularisation, using cross validation to select $\lambda$ [29].

### B. Experimental Setup

#### 1) Ellipticity detection

The catheter was inserted into a 3D printed tube with a circular cross-section of 25 mm diameter, Fig. 5B, and the balloon volume was increased then decreased in ten equal steps, with pressure and EIT measurements taken each time. Then, to simulate poor deployment of the balloon or outer stent, 3D printed inserts were placed within the circular tube creating an elliptical cross-section. Two sets of inserts were used with respective aspect ratios of $f = 0.75$ and $f = 0.5$. The test was repeated for both inserts, with the minor axis aligned with electrodes 1 and 5, and then rotated by 90° to align with 3 and 7. Firstly, the whole procedure was conducted using both the 8 measurement *radial* and the *full* EIT protocol.

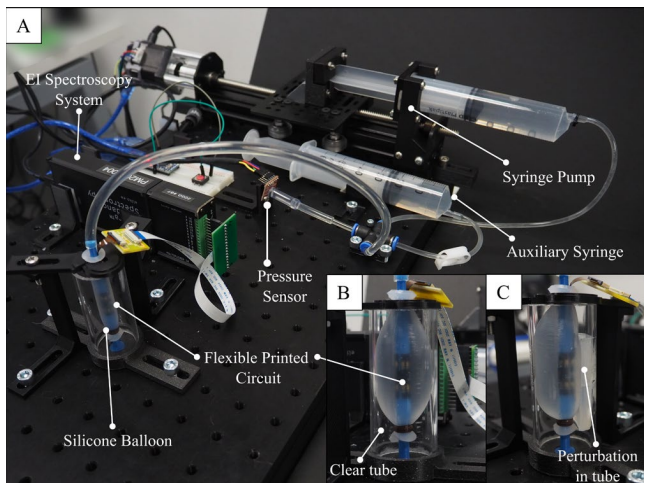

Fig 5. Hardware implementation of roboticised balloon catheter with impedance imaging (A) system overview (B) balloon catheter inflated in phantom lumen (C) lumen with occlusion

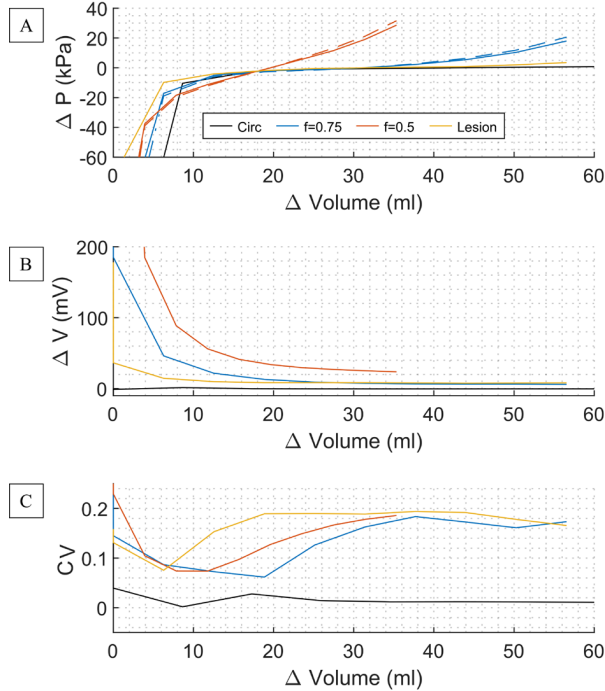

Fig 6. Ellipticity and lesion detection results for circular, $f = [0.75, 0.5]$ and lesion profiles (A) Pressure – volume curves, highlighting ambiguity between elliptical profiles rotated by 90° (dashed lines) (B) averaged change in voltages measured during inflation (C) coefficient of variation (CV) of measurements compared to expected values

*2) Lesion detection*

Experiments were then performed in a partially occluded lumen, designed to replicate the use case of localising stenosis or lesion within a vessel. A crescent-shaped silicone perturbation (Mold Star 19T, Smooth-On, USA), Fig. 5C, was attached inside a vertically supported 26 mm inner diameter Perspex tube. The data collection procedure was performed for a tube containing the lesion and for an empty tube with no lesion, for both *radial* and *full* EIT protocols.

*3) Lumen dilation*

The final experiments replicated the dilation of the lumen by the balloon inflation, such as the widening of the stenosis during angioplasty. The catheter was placed inside a silicone tube (Ecoflex 00-30, Smooth-On, USA) of inner diameter 24 mm that could itself deform when pressed by the inflated balloon. A second linear actuator was used to deform the silicone lumen and consequently the balloon underneath using a rubber-tipped rod. The tip was aligned with the centre of the FPC and extended into the lumen by 4 mm from the wall, which was used as the reference data frame. The tip was then retracted at a constant speed of 1 mm/s until it was no longer in contact with the lumen. EIT measurements were recorded continuously using only the *full* EIT protocol. To approximate the balloon cross section, a threshold was applied to the EIT images where elements with a δσ greater than three times the average value at the electrodes were removed.

## IV. RESULTS

*1) Ellipticity detection*

The initial increase in pressure during inflation, Fig. 6A, demonstrated the same trend for all experiments until the point of contact with the rigid wall, where the pressure sharply increased. The $f = 0.5$ lumen had a much steeper increase as the balloon met the rigid wall at lower volumes compared to $f = 0.75$. Rotating the elliptical profiles did not alter the pressure-volume curves (dashed lines), highlighting the difficulties in estimating shape from pressure measurement alone. The impedance measurements decreased with increasing balloon diameter, Fig. 6B, as expected from the container data Fig. 4D. Subsequently, the data was calibrated by subtracting data collected from the balloon inflating in free space. The coefficient of variation (CV) of the calibrated measurements, Fig 6C, demonstrates the deviations due to the occlusions. The increasing non-uniformity of the *radial* voltages resulting from the elliptical profiles is reflected in the CV. The differing elliptical profiles can also be observed in the voltages at $p_{max}$ in Fig 7, where the voltages increase at the minor axis as the wall is closer to the electrodes, and a decrease at the major axis as the balloon over inflates in that direction to compensate. The peak voltages shift by three electrodes when the lumen is rotated by 90° (dashed lines) which reflects the change in minor axis alignment.

The EIT images reconstructed using the full protocol reflect the same trend as the truncated radial data. The Abs and PTD reconstructions are uniform for circular cross section, and both show clear regions of conductivity decrease along the minor axis of the $f = 0.5$ ellipse, Fig. 8. However, a distinct profile was not present in the $f = 0.75$ images.

*2) Lesion detection*

The pressure curve, Fig. 6A, and the decrease in voltage, Fig. 6B was like those of the elliptical cross sections. The increase in CV Fig 6C shows the deviation after 10 mL, earlier than the pressure data. The lesion is readily identifiable in the calibrated measurements at $p_{max}$, Fig 7, with a single peak spread across more measurements compared to the narrower double peak in the elliptical data. These more identifiable voltage changes also corresponded to improved imaging. The *Abs* reconstruction located the lesion as a region of conductivity decrease, although the size was exaggerated and there were some artefactual changes on the opposing side, whereas *PTD* images had no such artefactual changes but further exaggerated the extent of the perturbation.

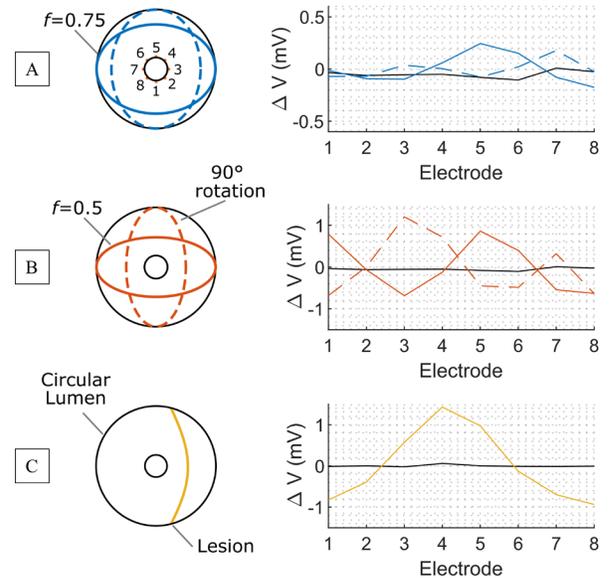

Fig 7. Ellipticity and lesion detection results: calibrated voltages at $p_{max}$ compared to measurements in unform circular lumen (*black*). Elliptical data with minor axis aligned with electrodes 3-7 *solid* and 1-5 *dashed* for $f = 0.75$ (A) and $f = 0.5$ (B), and (C) lesion aligned with electrodes 2-4

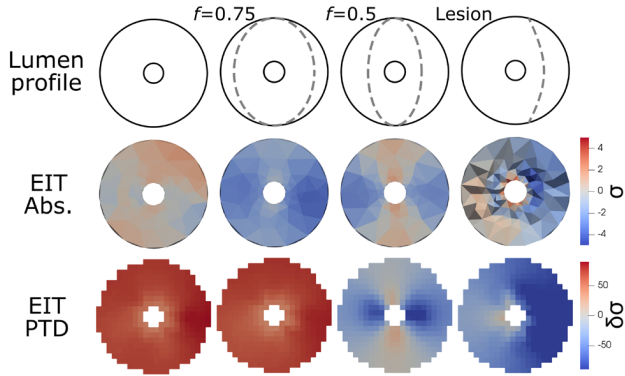

Fig 8. EIT image reconstructions of data collected in each lumen profile, using the *full* measurement protocol and nominal 30 mm mesh. *Abs:* Absolute reconstructions of conductivity σ without reference data, *PTD:* Pseudo Time Difference reconstruction δσ with simulated reference data

*3) Lumen dilation*

As the occlusion was removed from the lumen, there was a clear decrease in the voltages recorded as the balloon inflated to fill the volume, Fig. 9. A clear perturbation was reconstructed in the occluded region in both the *TD* and *PTD* images. Created without calibration, the *PTD* images demonstrate the extent of the change within the balloon, whereas the *TD* images relate to the extent of lumen dilation. The reconstructed cross section *CSA Approx.* better reflected the ground truth at 2 and 4 s when the occlusion was larger.

## V. DISCUSSION

The experimental results demonstrate the concept that balloon catheter sizing through impedance measurements can be extended to measure ellipticity and reconstruct images of occlusions. The results were in broad agreement with the detectability limits from numerical simulations, Fig. 3. For a 25 mm major axis, ratios below 0.75 were predicted to be detectable, which was reflected in the *radial* 8 channel data in Fig. 7 after calibration. However, images were not successfully reconstructed at this higher ratio, Fig 8. These results highlight the importance of careful calibration as the interchannel differences are greater than the 60 dB noise added in simulations. This noise could be reduced by replacing the ENIG coating on the FPC with either a manufacturing process with purer gold content or custom pre-treatment [30]. Imaging of the lesion Fig 7C and dilation of the lumen Fig 8 was more successful, as the conductivity contrast was localised to a single hemisphere of the balloon.

The advantage of the lack of reference frame in the *Absolute* solver came at the expense of significantly increased calculation time: 305 s compared to 62 ms for *PTD* on an Intel i7-7700, as it required 49 forward solves for 4 iterations.

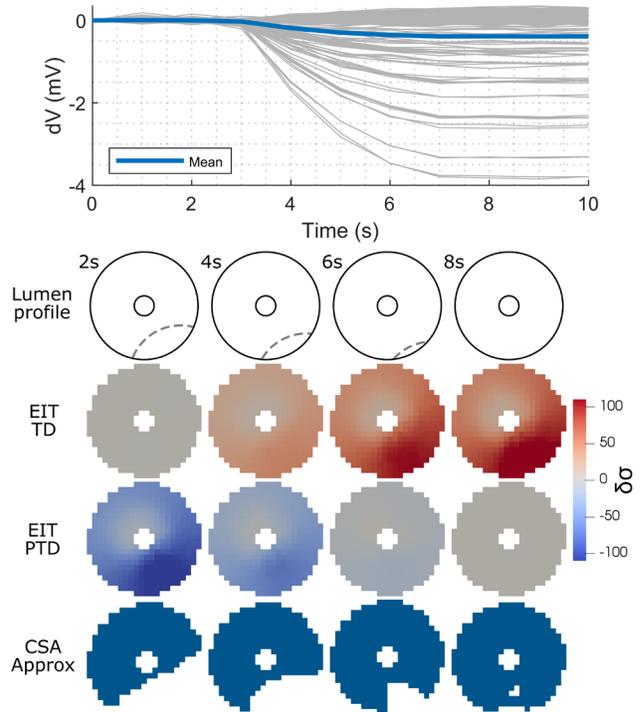

Fig 9. Lumen dilation results (Top) Voltage changes as balloon inflates as lumen is widened (Bottom) EIT reconstructions of changes δσ with respect to initial point *TD*, or simulated circular lumen *PTD*, lumen profile estimate from thresholding *PTD* image

Constraining the absolute solver to the two known conductivities may accelerate convergence and improve accuracy. As would replacing the NOSER [28] prior, which biases the reconstruction towards the electrodes and away from the target deformations at the balloon wall. Promising work combining absolute imaging with deep learning may be adaptable to internal electrode geometries to further improve reconstructions [31]. Zeroth order Tikhonov regularisation implemented in both *TD* algorithms provides stable reconstructions by assuming a smooth conductivity change [27]. Further improvements to the occlusion localisation may be possible through implementing regularisation which favours sharper or localised reconstructions, such as First order Tikhonov [32] or Total Variation [33]. EIT images could be potentially improved by using meshes closer to the nominal lumen diameter. This could be estimated from presurgical imaging [11] or by leveraging pressure/volume data to select an appropriate precalculated mesh. Combining EIT information with existing analytical models and pressure data [13] could further improve estimation.

Reconstructions were successful at a greater distance from the electrodes compared to other catheter based EIT imaging [18] due to the larger impedance changes caused by the balloon deformation. However, the disadvantage of this approach is that it cannot characterise the electrical properties of the tissue as the electrodes are isolated by the balloon. Increasing the number of electrodes could potentially further improve imaging by providing more subdivisions along the catheter surface or by creating a greater number of rings. The latter would provide a profile of the cross-sectional area and ellipticity, similar to that of provided by the FLIP [17]. Whilst the surface of the 14 Fr catheter could accommodate a far greater number of electrodes, the limitations arise from manufacturing the wiring along the length of the entire catheter. The 16 electrodes used here are of the order used in the FLIP, cochlea implants [34] and RF ablation catheters [35], and are thus feasible under the current state-of-the-art.

Changes in balloon diameter became harder to detect as the diameter increased, Fig. 3A, due to the reduction in sensitivity with distance from the electrodes, a common problem in EIT [27]. Adding electrodes to the inner surface of the balloon would greatly improve the sensitivity at the wall but presents a significant manufacturing challenge given the elongation of the material. Multifunctional stretchable sensors developed using thin film techniques have been successfully transferred to the balloon surface [36], although limited to smaller inflation ratios than used in this work. RF ablation balloon catheters currently on the market, have high inflation ratios and multiple electrodes [35]. By incorporating collapsible flexible traces on the balloon surface, the wiring does not need to stretch significantly, and thus a design employing similar techniques could increase sensitivity at the wall.

The silicone rubber balloon was inhomogeneous or not held in uniform tension and thus was not perfectly circular in the initial state and did not inflate entirely uniformly. This was visible in the change in voltages with the *radial* protocol Fig. 7 where even in a circular lumen, there was a greater change at electrode 6 and 7. These changes then render the variations caused by the elliptical profiles harder to detect. EIT sensitivity is greater closest to the electrodes, and thus even small variations at low inflations can cause changes larger than those measured at the lumen wall. Therefore, improving the uniformity of the balloon and the inflation are an important step to improve calibration. However, for stent deployment or angioplasty, the balloons are typically manufactured from polyurethane through blow moulding and in some cases the resulting stiffness requires inflation pressures over 10 times those recorded in the compliant balloon used in these experiments [5], [13]. This increased stiffness may benefit the calibration of the EIT measurements by improving the uniformity of the balloon prior to contact with the lumen.

Closed loop control of balloon diameter using the roboticised catheter, similar to that achieved through OCT [14], [15] is a logical extension of this proof of concept. This requires improvements to the communication between the EIT and syringe pump systems, to process the EIT data in real time. The frame rate of the Eliko Quadra system was ~1.5 Hz as each of the 136 impedance measurements were obtained sequentially. Changing to a parallel EIT system with multiple current sources could enable frame rates of > 50 Hz depending on the protocol required [29]. Imaging at this speed would also benefit from porting the reconstruction code to a compiled language such as Julia or C/C++.

The prototype described here is based on a 14 Fr catheter, but the approach can be generalised to other catheter sizes by following the numerical methods. However, decreasing the catheter diameter increases the difficulty in manufacturing the electrodes and locating them on the catheter sheath. The simulation, processing and imaging code is available at (https://github.com/EIT-team/Balloon_Catheter_Sizing).

## VI. CONCLUSIONS AND FURTHER WORK

This work presents a proof of concept of a novel approach to lumen sizing and occlusion imaging within a balloon catheter. EIT reconstructions of the inside of the balloon were obtained using multiple impedance measurements from electrodes on the central catheter shaft. The electrode arrangement was optimised in simulation and validated in phantom experiments. Experiments were then performed in three different use cases: ellipticity detection, lesion localisation and lumen dilation. Through calibration it was possible to detect aspect ratios of 0.75 and localise an occlusion inside the lumen. EIT imaging further successfully located the occlusion and visualised the dilation of the lumen.

Reducing interchannel variations by improved electrode coating and uniform balloon properties would aid calibration and further improve occlusion detection. Development of a reconstruction algorithm optimised to this geometry would further improve the localisation. These improvements could then be incorporated into an existing clinical balloon catheter for experiments in biologically representative phantoms.

## REFERENCES

[1] M. J. Kern, P. Sorajja, M. J. Lim, and M. J. Preceded by: Kern, "The interventional cardiac catheterization handbook."

[2] L. Robertson, K. I. Paraskevas, and M. Stewart, "Angioplasty and stenting for peripheral arterial disease of the lower limbs: An overview of Cochrane Reviews," *Cochrane Database Syst. Rev.*, vol. 2017, no. 2, 2017.

[3] G. Albert Schweitzer Hospital, Lambarene, Gabon, and Institute of Tropical Medicine, University of Tübingen, Tübingen, "Transcatheter versus Surgical Aortic-Valve Replacement in High-Risk Patients," *N. Engl. J. Med.*, vol. 365, pp. 687–696, 2011.

[4] M. C. Svendsen *et al.*, "Two-in-one aortic valve sizing and valvuloplasty conductance balloon catheter," *Catheter. Cardiovasc. Interv.*, vol. 86, no. 1, pp. 136–143, Jul. 2015.

[5] O. Fröbert, G. Sarno, S. K. James, N. Saleh, and B. Lagerqvist, "Effect of Stent Inflation Pressure and Post-Dilatation on the Outcome of Coronary Artery Intervention. A Report of More than 90 000 Stent Implantations," *PLoS One*, vol. 8, no. 2, pp. 1–10, 2013.

[6] T. H. Yang *et al.*, "Incidence and severity of paravalvular aortic regurgitation with multidetector computed tomography nominal area oversizing or undersizing after transcatheter heart valve replacement with the sapien 3: A comparison with the sapien XT," *JACC Cardiovasc. Interv.*, vol. 8, no. 3, pp. 462–471, 2015.

[7] S. K. Kodali *et al.*, "Two-Year Outcomes After Transcatheter or Surgical Aortic Valve Replacement," *Surv. Anesthesiol.*, vol. 57, no. 4, pp. 166–167, 2013.

[8] A. B. Freitas-Ferraz *et al.*, "Aortic Stenosis and Small Aortic Annulus: Clinical Challenges and Current Therapeutic Alternatives," *Circulation*, vol. 139, no. 23, pp. 2685–2702, 2019.

[9] R. Zegdi *et al.*, "Is It Reasonable to Treat All Calcified Stenotic Aortic Valves With a Valved Stent?. Results From a Human Anatomic Study in Adults," *J. Am. Coll. Cardiol.*, vol. 51, no. 5, pp. 579–584, 2008.

[10] E. Altiok *et al.*, "Comparison of two-dimensional and three-dimensional imaging techniques for measurement of aortic annulus diameters before transcatheter aortic valve implantation," *Heart*, vol. 97, no. 19, pp. 1578–1584, 2011.

[11] A. M. Kasel *et al.*, "Standardized imaging for aortic annular sizing: Implications for transcatheter valve selection," *JACC Cardiovasc. Imaging*, vol. 6, no. 2, pp. 249–262, 2013.

[12] D. Leibowitz *et al.*, "Computerized gradual balloon inflation: a novel strategy of coronary angioplasty superior to a standard manual approach," *Cardiovasc. Revascularization Med.*, vol. 10, no. 1, pp. 45–48, 2009.

[13] A. Palombi *et al.*, "Sizing the aortic annulus with a robotised,

commercially available soft balloon catheter: In vitro study on idealised phantoms," *Proc. - IEEE Int. Conf. Robot. Autom.*, vol. 2019-May, pp. 6230–6236, 2019.

[14] H. Azarnoush, S. Vergnole, B. Boulet, M. Sowa, and G. Lamouche, "Real-time control of angioplasty balloon inflation based on feedback from intravascular optical coherence tomography: Experimental validation on an excised heart and a beating heart model," *IEEE Trans. Biomed. Eng.*, vol. 59, no. 5, pp. 1488–1495, 2012.

[15] H. Azarnoush, S. Vergnole, V. Pazos, C.-É. Bisaillon, B. Boulet, and G. Lamouche, "Intravascular optical coherence tomography to characterize tissue deformation during angioplasty: preliminary experiments with artery phantoms," *J. Biomed. Opt.*, vol. 17, no. 9, p. 1, 2012.

[16] M. C. Svendsen *et al.*, "Conductance sizing balloon for measurement of peripheral artery minimal stent area," *J. Vasc. Surg.*, vol. 60, no. 3, pp. 759–766, 2014.

[17] Z. Lin *et al.*, "Functional luminal imaging probe topography: an improved method for characterizing esophageal distensibility in eosinophilic esophagitis," *Therap. Adv. Gastroenterol.*, vol. 6, no. 2, pp. 97–107, Mar. 2013.

[18] Y. Luo, D. Huang, Z.-Y. Huang, T. K. Hsiai, and Y.-C. Tai, "An *ex vivo* Study of Outward Electrical Impedance Tomography (OEIT) for Intravascular Imaging," *IEEE Trans. Biomed. Eng.*, vol. 9294, no. c, pp. 1–1, 2021.

[19] T. Zhao, C. Wu, and M. Soleimani, "Ionic liquid based distributed touch sensor using electrical impedance tomography," *IOP SciNotes*, vol. 1, no. 2, p. 025005, 2020.

[20] J. Avery, D. Shulakova, M. Runciman, G. P. Mylonas, and A. Darzi, "Tactile Sensor for Minimally Invasive Surgery Using Electrical Impedance Tomography," *IEEE Trans. Med. Robot. Bionics*, vol. 2, no. 4, pp. 561–564, Nov. 2020.

[21] S. Russo, T. Ranzani, H. Liu, S. Nefti-Meziani, K. Althoefer, and A. Menciassi, "Soft and stretchable sensor using biocompatible electrodes and liquid for medical applications," *Soft Robot.*, vol. 2, no. 4, pp. 146–154, 2015.

[22] J.-B. Chossat, H.-S. Shin, Y.-L. Park, and V. Duchaine, "Soft Tactile Skin Using an Embedded Ionic Liquid and Tomographic Imaging," *J. Mech. Robot.*, vol. 7, no. 2, p. 021008, May 2015.

[23] P. C. Patsalis *et al.*, "Undersizing but overfilling eliminates the gray zones of sizing for transcatheter aortic valve replacement with the balloon-expandable bioprosthesis," *IJC Hear. Vasc.*, vol. 30, p. 100593, Oct. 2020.

[24] A. Adler and W. R. B. Lionheart, "Uses and abuses of EIDORS: an extensible software base for EIT," *Physiol. Meas.*, vol. 27, no. 5, pp. S25–S42, May 2006.

[25] M. Min, M. Lehti-Polojärvi, P. Annus, M. Rist, R. Land, and J. Hyttinen, "Bioimpedance spectro-tomography system using binary multifrequency excitation," *11th Int. Conf. Bioelectromagn.*, no. May, pp. 5–8, 2018.

[26] V. Kolehmainen, M. Vauhkonen, P. a Karjalainen, and J. P. Kaipio, "Assessment of errors in static electrical impedance tomography with adjacent and trigonometric current patterns.," *Physiol. Meas.*, vol. 18, no. 4, pp. 289–303, 1997.

[27] D. S. Holder, "Electrical Impedance Tomography: Methods, History and Applications," *CRC Press*, p. 456, 2004.

[28] M. Cheney, D. Isaacson, J. C. Newell, S. Simske, and J. Goble, "NOSER: An algorithm for solving the inverse conductivity problem," *Int. J. Imaging Syst. Technol.*, vol. 2, no. 2, pp. 66–75, 1990.

[29] J. Avery, M. Runciman, A. Darzi, and G. P. Mylonas, "Shape Sensing of Variable Stiffness Soft Robots using Electrical Impedance Tomography," in *2019 International Conference on Robotics and Automation (ICRA)*, 2019, pp. 9066–9072.

[30] S. Anastasova, P. Kassanos, and G. Z. Yang, "Multi-parametric rigid and flexible, low-cost, disposable sensing platforms for biomedical applications," *Biosens. Bioelectron.*, vol. 102, no. October 2017, pp. 668–675, 2018.

[31] S. J. Hamilton, A. Hänninen, A. Hauptmann, and V. Kolehmainen, "Beltrami-net: Domain-independent deep D-bar learning for absolute imaging with electrical impedance tomography (a-EIT)," *Physiol. Meas.*, vol. 40, no. 7, 2019.

[32] M. Jehl, J. Avery, E. Malone, D. Holder, and T. Betcke, "Correcting electrode modelling errors in EIT on realistic 3D head models," *Physiol. Meas.*, vol. 36, no. 12, pp. 2423–2442, Dec. 2015.

[33] J. Liu, L. Lin, W. Zhang, and G. Li, "A novel combined regularization algorithm of total variation and Tikhonov regularization for open electrical impedance tomography," *Physiol. Meas.*, vol. 34, no. 7, 2013.

[34] A. Dhanasingh and C. Jolly, "An overview of cochlear implant electrode array designs," *Hear. Res.*, vol. 356, pp. 93–103, 2017.

[35] G. S. Dhillon *et al.*, "Use of a multi-electrode radiofrequency balloon catheter to achieve pulmonary vein isolation in patients with paroxysmal atrial fibrillation: 12-Month outcomes of the RADIANCE study," *J. Cardiovasc. Electrophysiol.*, vol. 31, no. 6, pp. 1259–1269, 2020.

[36] D. Kim *et al.*, "Materials for Multifunctional Balloon Catheters With Capabilities," *Nat. Mater.*, vol. 10, no. 4, pp. 316–323, 2011.